\documentclass[preprint,12pt]{elsarticle}

\usepackage{etoolbox}
\patchcmd{\abstract}{Abstract}{Abstract}{}{}

\usepackage{float}
\usepackage{graphicx}
\usepackage{amssymb}
\usepackage{amsmath}
\usepackage{lineno}
\journal{Arxiv}

\begin{document}

\begin{frontmatter}

\title{Small Moving Window Calibration Models for Soft Sensing Processes with Limited History}
%\title{Hybrid Partial Least Squares Random Forest Regression Ensemble as a Soft Sensor}

\author{Casey Kneale}
\author{Steven D. Brown\corref{cor1}}
\cortext[cor1]{sdb@udel.edu}
\address{Department of Chemistry and Biochemistry, University of Delaware, 163 the Green, Newark, DE, 19716, USA\fnref{label3}}


%%\author{Casey Kneale}
%%\author{Steven D. Brown}
%%address{Delaware, United States}

\begin{abstract}
Five simple soft sensor methodologies with two update conditions were compared on two experimentally-obtained datasets and one simulated dataset. The soft sensors investigated were moving window partial least squares regression (and a recursive variant), moving window random forest regression, the mean moving window of $y$, and a novel random forest partial least squares regression ensemble (RF-PLS), all of which can be used with small sample sizes so that they can be rapidly placed online. It was found that, on two of the datasets studied, small window sizes led to the lowest prediction errors for all of the moving window methods studied. On the majority of datasets studied, the RF-PLS calibration method offered the lowest one-step-ahead prediction errors compared to those of the other methods, and it demonstrated greater predictive stability at larger time delays than moving window PLS alone. It was found that both the random forest and RF-PLS methods most adequately modeled the datasets that did not feature purely monotonic increases in property values, but that both methods performed more poorly than moving window PLS models on one dataset with purely monotonic property values. Other data dependent findings are presented and discussed.
\end{abstract}


\end{frontmatter}

%\linenumbers

\section{Introduction}
\label{S:1}
Soft sensors for regression tasks have found wide utility in process engineering and process analytical chemistry \cite{openingclaim, kadlec, GoodReview}. A soft sensor is effectively a calibration used on time-series data. Here, we consider a soft sensor to be any algorithm that can be used to estimate a property value from several readily available but indirect measurements. The goal of implementing a soft sensor is typically to avoid the use of a physical sensor for variables that may require extensive time or work up to measure \cite{GoodReview}. In the context of industrial chemical processes, these algorithms should meet several specifications. In most situations, soft sensors must process multivariate data quickly and should also predict a property of interest despite delayed or infrequent laboratory reference measurements \cite{GoodReview}.

There are several general approaches to soft sensing that satisfy these requirements. A simple approach is to build a global process model from off-line historical data and to apply that model to online data \cite{NNBookbyQin,datarichinfopoor}. This method has been shown to work if a process is stable \cite{NNBookbyQin}, but real processes can feature unpredictable phenomena or subtle process changes (sensor drift, sensor replacement, etc), that may not have been encountered in the historical data and are therefore not included in the historical model. These unanticipated effects in the process can render a soft sensing model inaccurate \cite{lessonslearned}. The collection of a historical data bank can also be costly and time consuming for new processes. 

A less data-intensive but common approach to soft sensing is to use local regression models \cite{local} built from on-line and historical data. Local models can be described as a collection of several calibration models that are individually assigned to data collected over a specific region in time, such that each calibration corresponds to only a portion of the process, sometimes called a state in the data. Several notable examples of process state identification and calibration include ones based on finite mixtures of Gaussian models \cite{FGMM} and mixtures of partial least squares ensembles \cite{mixPLS}. These local modeling methods have been shown to work well, but they often involve complicated implementations, and lead to relatively large, complex models. In addition, when new process states are found, the model must be put offline to complete a new calibration.

Considerable research has been devoted to use of dynamically or adaptively updated models to resolve problems with local modeling methods. Some dynamic soft sensing strategies adapt the inner and outer relations of partial least squares (PLS) regression models as new information becomes available through various assumptions and statistical mechanisms \cite{dynamicPLS, dynamicPLS2}. One of the simpler and more widely used methods for adaptive soft sensing is recursive partial least squares (RPLS)  regression \cite{QinRPLS,Dayal}. These recursive methods allow real-time updates of the PLS model with information from recently acquired samples at the expense of tuning a forgetting factor \cite{QinRPLS}. To overcome issues that arise with nonspecific dynamic updates, methods which combine local modeling with adaptive updates \cite{local, forsdb} have also been explored. Unfortunately, like many local process models, most of these dynamic modeling methods have complex implementations, require many assumptions about the data, involve significant model tuning \cite{FIRPLS}, and/or require a sizable bank of historical data to initialize the calibration.

The purpose of this study was to assess the performance of models for new processes given limited amounts of historical data. The primary goal of this investigation was to study how well simple regression models calibrate process time series data using few training samples, so that they can be readily placed online. These methods may find use in the application to new processes, or during the time where new process states occur in a modeled processes and a more complicated model is put offline. A secondary goal of the study was that the methods featured here should be easy to implement and require little tuning to expedite their use.

A moving window modeling approach is one simple way to make regression methods dynamic, so that they can be used with process data. A moving window model can be easily implemented \cite{rollingwindow}, is fast to compute, does not rely on process mode identification, and it can have short start-up times to allow for its application on new processes with unknown operating characteristics. However, a moving window modeling approach such as that used in this study requires that laboratory reference measurements are updated throughout the process. 

In this work, we present and compare the predictive accuracy of five simple soft sensing strategies for regression tasks on multivariate chemical process data, using regression models developed on small moving windows. The widely used mean moving window \cite{rollingwindow}, moving window partial least squares (PLS) \cite{StateOfTheArt}, and moving window updated recursive partial least squares (RPLS) \cite{QinRPLS}, were compared with moving window random forest regression (RF), and a novel, random forest-partial least squares regression ensemble (RF-PLS) method. These modeling approaches can be used in real-time applications for one or several-time-step ahead predictions for soft sensing with little a priori knowledge and with minimal assumptions about the data.

\section{Theory}
\subsection{Soft sensing}
Soft sensing is a term used to describe a form of multivariate calibration that relates property values to indirect measurements made on a multivariate time series. Soft sensors are typically models based on physical measurements (temperature, pressure, etc) made on a process to infer chemical properties (concentration, purity, activity, etc). The overall goal of soft sensing is to predict chemical properties indirectly from measurements made on physical sensors. The aim is to obtain values for properties more rapidly or economically than a direct measurement of those properties \cite{openingclaim, kadlec, GoodReview}. Because of the similarities between soft-sensing and multivariate calibration, standard modeling techniques, such as partial least squares regression, can be used for soft sensing, though these techniques require some adjustments to deal with time dependent data from a process \cite{local}.

\subsection{Partial Least Squares Regression}
Partial least squares regression is a regression technique commonly employed in chemometrics \cite{PLSreg}. Like traditional least-squares regression, dependent observations ($y$) are regressed onto the independent data ($X$) using the standard linear form, $y = X\hat{B} + e_y$, where $\hat{B}$ represents the regression coefficients, and $e_y$ is the residual error from the fit to the corresponding true $y$ vector. In PLS regression, unlike multiple linear regression, the regression coefficients are constructed from a bilinear decomposition of the covariance between $y$ and $X$,
\begin{equation}
	\begin{aligned}
		y = Uq^T + e_Y \\
	\end{aligned}
\end{equation}
\begin{equation}
	\begin{aligned}
		X = TP^T + E_X
	\end{aligned}
\end{equation}

\noindent where $U$ and $T$ are the projections (scores) from $y$ and $X$ onto the model space. Equations 1 and 2 represent the so-called outer relations in PLS; they relate the $X$ and $y$ blocks to latent features. PLS regression can be used to perform a decomposition of the $X^Tyy^TX$ matrix such that the covariance between $T$ and $U$ in the outer relations is maximized, which allows for an inner relationship between the $X$ and $y$ blocks to be exploited. The estimated regression coefficients $\hat{B}$ are calculated from both equations via the weights ($W$), of the least squares projections from $T$ to $P$ and the $y$ loadings, \begin{equation} \hat{B} = Wq^T \end{equation}

The PLS method overcomes collinearity in $X$ data and allows for dimension reduction to latent variables. These two features provide a notable advantage over many other regression methods and because of this capability, PLS has been shown to be well-suited for many regression tasks in chemometrics. However, for time-dependent data such as those from chemical processes, the covariance of $X$ is not fixed; it often changes over time and in unpredictable ways. Therefore, the usual static calibration models developed from PLS are not suitable for soft-sensing.

\subsection{Moving Window Calibrations}
One simple adaptation to make multivariate calibrations more responsive to changes in the autocovariance of the time-dependent data in $X$, is to employ a moving window. The use of the term ``moving window" in this work follows that of Qin \cite{QinRPLS}; this is equivalent to the term rolling window which is commonly used in the statistical literature \cite{rollingwindow}. This moving window modeling approach for time-dependent data differs considerably from the variable selection technique built on time independent $X$ data, sometimes called moving window PLS (MWPLS)\cite{varselect} in the chemometrics literature.

A window is a collection of adjacent samples in the process with known property values that is selected for calibration. Many variations on this approach have been reported \cite{QinRPLS,MWvariant}, but only the simplest case, that of modeling time-series data with a moving window, is examined here. A local calibration model built from the window of time-dependent data may be used to predict properties of samples collected at future time steps. To account for changes in the covariance of time-dependent data, a calibration window of fixed sample length can then be moved one or more time steps ahead, and provided that reference measurements exist a new calibration model can be trained on those samples now in the calibration window. Again, the local model can then be used to predict property values for future data. When new property values become available, the calibration window can then be updated again. This sequence of modeling and prediction can be repeated for any number of incoming samples. We refer to the number of sample time steps from the last training example in the moving calibration window to the unknown measurements as the delay.  

We refer to the case where the modeling window is moved forward by one time step after every prediction as the ``continuously updated case" because the calibration model is updated for each newly available sample. This case presumes frequent measurements of property values and rapid availability of the data. In a similar way, we refer to the ``delayed update case" as the case where the calibration window is updated less frequently than at every time step. This case is appropriate where there is a delay in making property values available to the model. These two cases are shown in Figure 1. We applied both update methodologies to every regression method tested in this study, as described in the methods section.

\begin{figure}[H]
	\centering
	\includegraphics[width=0.8\linewidth]{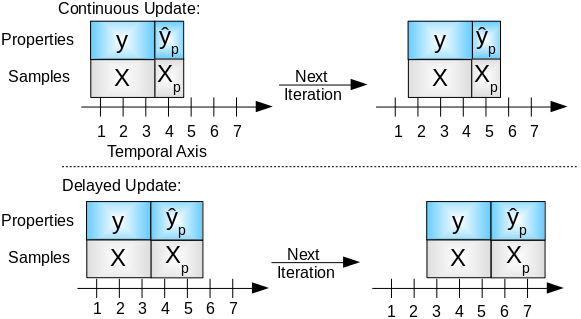}
	\caption{Illustrative depictions of models which are continuously updated (above) and those which are updated with a delay (lower). The horizontal axis depicts samples and their temporal order, ``X" labeled boxes indicate the observations used to construct models, ``y" labeled boxes are the property values used for calibration, and ``X$_p$" or ``$\hat{y}_p$" labeled boxes are the observations and predictions which are made from the model built on samples within the calibration window. In the continuous case, the calibration window moves with a fixed delay relative to the sample for prediction; in this example the calibration window is one time step behind the prediction sample. In the delayed update case, several samples are predicted from one window location before the calibration window is moved and updated. In this example, the window size is three time steps, and the model developed from data in the window predicts two new samples before it is updated.}
\end{figure}

Although the moving window approach can be used to overcome time-related changes in the autocovariance, this approach does not implicitly permit the regression model developed for the window of data to retain information from samples collected prior to the time spanning the sample window. 

\subsection{Recursive PLS}
One adaptation to PLS regression that allows modeling of time-series data is to make its model inputs recursive. Recursive model adaptation allows information from data collected at earlier times to be used in developing the model from the data in the current sample window. This idea was first pioneered by Helland, et al. \cite{HellandRPLS}.

Qin et. al. \cite{QinRPLS} showed that PLS algorithms which normalize the $X$ score vectors ($T$) but not the weights or $X$ loadings ($P$) can be made recursive. In the recursive algorithm, the $X$ and $y$ input matrices are modified to incorporate information from the previous PLS model and newly obtained data ($X_{new}$, $y_{new}$),

\begin{equation}
X = \begin{bmatrix} $$\lambda P_{old}^T$$ \\ $$X_{new}$$ \end{bmatrix} \qquad 
y = \begin{bmatrix} $$\lambda B_{old}Q_{old}^T$$ \\ $$y_{new}$$ \end{bmatrix} 
\end{equation}

\noindent where $B_{old}$ is a diagonal matrix of inner model coefficients which are defined for each latent variable as $u^T \cdot t$, and where $\lambda$ is a scalar forgetting factor. The forgetting factor determines the relative amount of information retained from previous models in the recursive update, allowing for new information to be incorporated into a weighted PLS model in an adaptive manner. 

Alternative methods to Qin's recursive algorithm, the one used in this study for PLS modeling, have also been proposed. For example, Dayal and McGregor \cite{Dayal} presented an exponentially weighted covariance matrix update method which is based on a kernel PLS regression algorithm. 

\subsection{Regression Trees}
Because random forest regression, a nonlinear decision tree ensemble model, has little precedence in soft sensing we include a brief overview of the algorithm's operating characteristics, weaknesses, and what motivates its use in the novel random forest-partial least squares regression method. For an overview of regression trees, refer to the work done by Breiman, et al. \cite{CART}, or that of Loh \cite{CARTez}.

Nonlinear regression techniques are thought to be useful for soft sensing because they can model the nonlinear relationships that are believed to be common in process data \cite{nonlinclaim}. One of the simplest nonlinear calibration methods is that of a regression tree. Breiman's classification and regression tree algorithm (CART) \cite{CART} is well known for classification, but its suitability for nonlinear regression is less often mentioned. CART trees are widely used for their simple interpretations and their ability to model nonlinearities in data \cite{CARTez}. 

Regression trees are collections of sequential, binary decisions, where each decision pertains to a selected independent variable. For example, assume that the matrix $X_{o,c}$ contains $o$ row-wise observations and $c$ columns or variables. A regression tree model is produced by selecting a variable $v$ from $[1,c]$, and two observations, $r_1$ and $r_2$, in $o$ to create a decision boundary from the mean of $r_1$ and $r_2$. The decision boundaries are then used to partition the remaining samples into two groups, depending on whether each sample is greater or less than $\frac{r_1 + r_2} {2}$. Each of these groups is considered a branch if further binary decisions partition the present data, or a node if the data are no longer partitioned after a decision boundary.

The variables and observations for each partition which create the decision boundary are selected such that a gain metric is optimized. The gain metric commonly used for regression trees is the sum of squared errors for prediction, defined as $SSE = \sum(y_{predicted} - y_{actual})^2$ \cite{CARTez}. This gain metric is unlike the gain metrics used in classification trees \cite{CART} in that it assesses regression accuracy via the sum of squared deviations of each terminal node rather than on class purity. The process of sample partitioning is then repeated in a recursive manner on each partition until a stopping condition is met. The most common stopping criterion for regression trees is reached either when the act of introducing a new decision or partition no longer organizes the samples into distinct groups which have similar property values, or when the number of samples in a given partition is less than a user-specified value \cite{randomforest}. 

Predicting from regression trees is easily accomplished after the tree has been constructed. New samples are passed through the hierarchy of binary decisions until the samples are fully described by the set of binary decisions created from the training samples. At this point, the samples are assumed to possess the same property value as the mean of the property values that populated this terminal node of the regression tree during training. Predicting from decision trees is computationally efficient relative to many other methods because only the selected variables, decision boundaries, and the property values at each terminal node need to be stored \cite{CARTez}. 

Although regression tree models are typically fast to train \cite{CARTez}, there are problems with the use of a single regression tree for a complex regression task. Decision boundaries constructed from minimizing the sum of squared errors of prediction tend to grow overly complex trees. Every unique instance of a property value will tend to result in its own terminal node. Thus, the decisions made on future data will be over-fit to the training data and the predictions from a CART tree are limited to discrete instances of property values from the training set.  

\subsection{Random Forest Regression}
The type of prediction formed by regression trees is in conflict with the notion that regressions should be able to interpolate between property values on which the model was trained. There are many schemes to overcome this issue, each with their own strengths and weaknesses. A notable example is the implementation of piecewise least squares regression modeling performed at tree nodes to allow for interpolation between training examples \cite{M5}. An alternative approach, the one taken in this work, is to grow many trees, each made from random selections of variables, and drawing randomly from a set of training samples contained in a moving window of process data taken over time with replacement. The average of the predictions from many trees with these stochastically-selected variables and samples tends to be a reliable estimate of the property value \cite{randomforest}. Most importantly, the average of many predictions from randomly constructed trees typically lies on a continuum and the predictions from a random forest model are not limited to the set of training samples like regression trees. This process of randomly sampling observations with replacement to build a model and averaging the predictive results is called bootstrap aggregating or bagging \cite{bagging}.

The technique of combining predictions from many regression trees via the use of bagging is referred to as random forest regression \cite{randomforest}. We utilized random forest regression on a moving window, as presented in the moving window calibration section, for soft sensing. Much like CART, Breiman's random forest algorithm is widely known for classification for its ability to robustly handle noise and to model nonlinearities in data \cite{nonlinear}, but its utility for nonlinear regression is less often mentioned. Random forest regression is not commonly used for soft sensing, but a limited study suggested that the performance of a random forest ensemble model was similar to that of a neural network for determining waste water quality from processing plant sensors \cite{wastewater}.  

\subsection{Random Forest-Partial Least Squares Regression Ensemble Modeling }
For small window sizes, like those studied here, it is common that the actual property value of an unknown exceeds the bounds of the training set. Even though random forest modeling can be used to interpolate between training sample property values on new data, an advantage over modeling with a single regression tree, the random forest regression cannot be used to predict property values outside of the range of the calibration. In cases where data have either a local or global monotonic trend, the predictions from random forest will therefore be biased to the range of the training examples. It should be stated that, in typical multivariate calibrations, the act of predicting outside of the range of a models training set is not a desirable practice. In time series, however, such predictions can be unavoidable. In practice we have found that random forest regression models made from small windows of samples which also incorporated information from PLS predictions can allow the random forest regression model to extrapolate predictions outside of its training range.

To extrapolate predictions outside of the values contained in a training set, as is possible in PLS regression, while also incorporating an ability to handle nonlinearities in changes across the sample covariance, such as that done in RF regression, a novel combination of both methods was created. Random forest partial least squares ensemble models are made by incorporating PLS prediction(s) of the unknown sample, as well as any measured reference samples which are typically included in a moving window model, into a single random forest regression model. Mathematically the prediction function for this hybrid method can be described as the following,
\begin{equation}
y_{t} = f{(x_{t},x_{t-1} , ... , x_{t-w}, \hat{y}^{PLS}_t, y_{t-1} , ... , y_{t-w})}
\end{equation}
where $\hat{y}^{PLS}_t$ is obtained via a moving window prediction from a conventional PLS regression model and $w$ is the number of samples included in the random forest regression window. With this adjustment to random forest regression, the bootstrap aggregating that is applied to the inner CART models then randomly incorporates the PLS model information into an otherwise typical RF model. This combination is shown diagrammatically in Figure 2. 

\begin{figure}[H]
	\centering
	\includegraphics[width=0.75\linewidth]{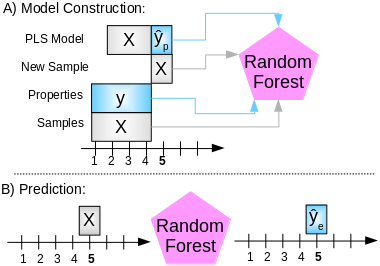}
	\caption{ (A) A depiction of how the random forest partial least squares regression ensemble is constructed from the data in $X$ and uses a conventional PLS prediction ($\hat{y}_p$) from a different window of samples to train the model. PLS models are created from samples (2-4) collected prior to the unknown (5), and their predictions of the unknown are used along samples with the known property values (1-4) to construct a random forest regression model. (B) The random forest ensemble model is then applied to the sensor information from the unknown sample (5) to predict its property value ($\hat{y}_e$.)}
\end{figure}

The introduction of PLS predictions allows for the linear prediction from one or more PLS models to extend the range of prediction in a random forest regression model. In our experiments, only one PLS prediction was utilized. Predicting from the random forest partial least squares ensemble model is performed by applying the $X$ information from the unknown sample to the random forest model. The hybrid model can be updated to predict upon new samples in an analogous way to that done in updating the moving window in a conventional regression model used for soft sensing. Both the PLS and random forest sample windows are reconstructed from samples which are one or more time steps ahead of the samples that were used to create both the interior PLS and exterior RF regression models.

Information leakage, as occurs upon training on test data, does not occur with proper implementation of this algorithm. Although the random forest regression models are constructed from PLS predictions onto the unknown sample, the interior PLS models do not contain the actual $y$ value, only one or more moving window-based PLS predictions of it. 

\subsection{Mean Moving Window}
The mean of the moving window (MMW), sometimes called the simple moving average\cite{rollingwindow}, was implemented in this study as a negative experimental control. The mean moving window model assumes that an unknown property value is the average property value of the window of the $w$ samples which proceeded it, 
\begin{equation}
y_{t} = {(y_{t-1}, y_{t-2}, ... , y_{t-w})}/w
\end{equation} The mean prediction of a moving window was implemented as a control because an estimate from the mean moving window assumes that all future predictions are expected to resemble their $w$ neighboring samples, and makes no use of any process information in $X$. An effective moving window soft sensor model should outperform the mean model if it incorporates information presented in $X$, in time-dependent changes in $y$, or in how $y$ changes with respect to changes in $X$.

\section{Data}
\subsection{Debutanizer process}
The debutanizer process data were collected from a distillation column which was designed to remove butane from gasoline. The goal of the process modeling was to quantify the amount of butane released from gasoline during the refining process. Butane amounts were inferred from seven measurements of temperature at the top of the column($v_1$), at the bottom of the column($v_6$, $v_7$), in the sixth tray($v_5$), in the redux flow ($v_3$) and in the flow to next process($v_4$), and from pressure at the top of the column ($v_2$) \cite{SSBook}. A total of 2394 samples were available. 

The sensor measurements were collected at a fixed time interval of 6 minutes, but the reference measurements took considerably longer to collect, 45 minutes. The property values for this dataset lagged the process variable measurments by 8 samples because of the time used to perform the butane quantitation. To align model predictions from the sensor data and the property values at the appropriate time, the property values were offset by 48 minutes. 

\subsection{Sulfur Recovery Unit Process}
The Sulfur Recovery Unit (SRU) process was designed to neutralize H$_2$S and SO$_2$ gasses and to retain elemental sulfur from furnaces in a chemical plant. The process utilized 5 gas flow sensors that collected measurements every minute, so that the concentrations of both gasses could be modeled \cite{SSBook}. The SRU dataset has many instances where consecutive property values for each gas have exactly the same numerical value. An artificial floating point precision offset ($1$x$10^{-6}$) was added to every property value that was identical to its earlier neighbor to avoid zero-variance data, which would otherwise cause partial least squares regression modeling to fail when applied to data collected over these short process windows.

\subsection{Simulated Penicillin Fermentation}
We also examined data from a simulated penicillin fermentation process. The process data consisted of 15 variables: aeration rate, agitator power, substrate feed rate, substrate feed temperature, acid flow rate, base flow rate, cooling water flow rate, substrate concentration, dissolved oxygen concentration, biomass concentration, penicillin concentration, culture volume, carbon dioxide concentration, pH, fermenter temperature, and generated heat\cite{pensim}. The goal was to determine the penicillin concentrations from the other process variables. With the exception of the sampling interval (1.1 hrs) and simulation time (350 hrs), all of the settings that were utilized in the simulation were the same as the published, default settings. 

\section{Methods}
Conventional partial least squares regression and random forest regression were the moving-window-based modeling methods selected for this study. Both PLS and RF regression used only information from $X$ and $y$ in the moving window of $w$ samples for calibration and had the following functional form for prediction, 
\begin{equation}
y_{t} = f{(x_{t},x_{t-1} , ... , x_{t-w}, y_{t-1} , ... , y_{t-w})}
\end{equation}
All PLS regression models were cross validated inside of each moving window by leave-one-out cross validation. Window sizes for all of the moving window models were validated by a train-test split by using the first 4.0\% of the debutanizer process data, 4.6\% of the sulfur recovery unit process data, and 14.1\% of the simulated penicillin data. In these studies the number of trees used in the random forest model was fixed at 1000 for reproducibility.

Recursive PLS regression was selected as the adaptive method to be compared with the other moving window techniques, so that results from a commonly used soft sensor modeling techniques could be compared with those from the random forest partial least squares ensemble. The lowest cross-validated root mean squared error of prediction (RMSEP) on the window sizes selected for this study were obtained from the RPLS modeling method on the debutanizer, sulfur recovery unit, and penicillin datasets utilized forgetting factors of 0.01, 0.05, 0.05, and 0.10, respectively.

Both of the previously mentioned updating approaches, the continuously updated case, and the delayed update case with varying update frequencies, were investigated for each soft sensor modeling technique and for each dataset. The window sizes used for the models were determined empirically, as described below.

\section{Results and Discussion}

\subsection{Effect of Window Size on One-Step-Ahead Predictions}
An experiment to determine the relationship, if any, between the one-step-ahead predictive error and moving window size was performed on the debutanizer and sulfur recovery unit datasets. Figure 3 shows the one-step-ahead prediction errors obtained by using MMW, PLS, RPLS, and RF models, which each had window sizes of 2-10, 15, 20, and 25 samples. It was found that across a given dataset, as the size of the process model window increased, so did the prediction error. The trend between window size and one-step-ahead prediction error was similar to what was observed during train-test split validation as can be seen in the supplementary information. 

\begin{figure}[H]
	\centering
	\includegraphics[width=0.9\linewidth]{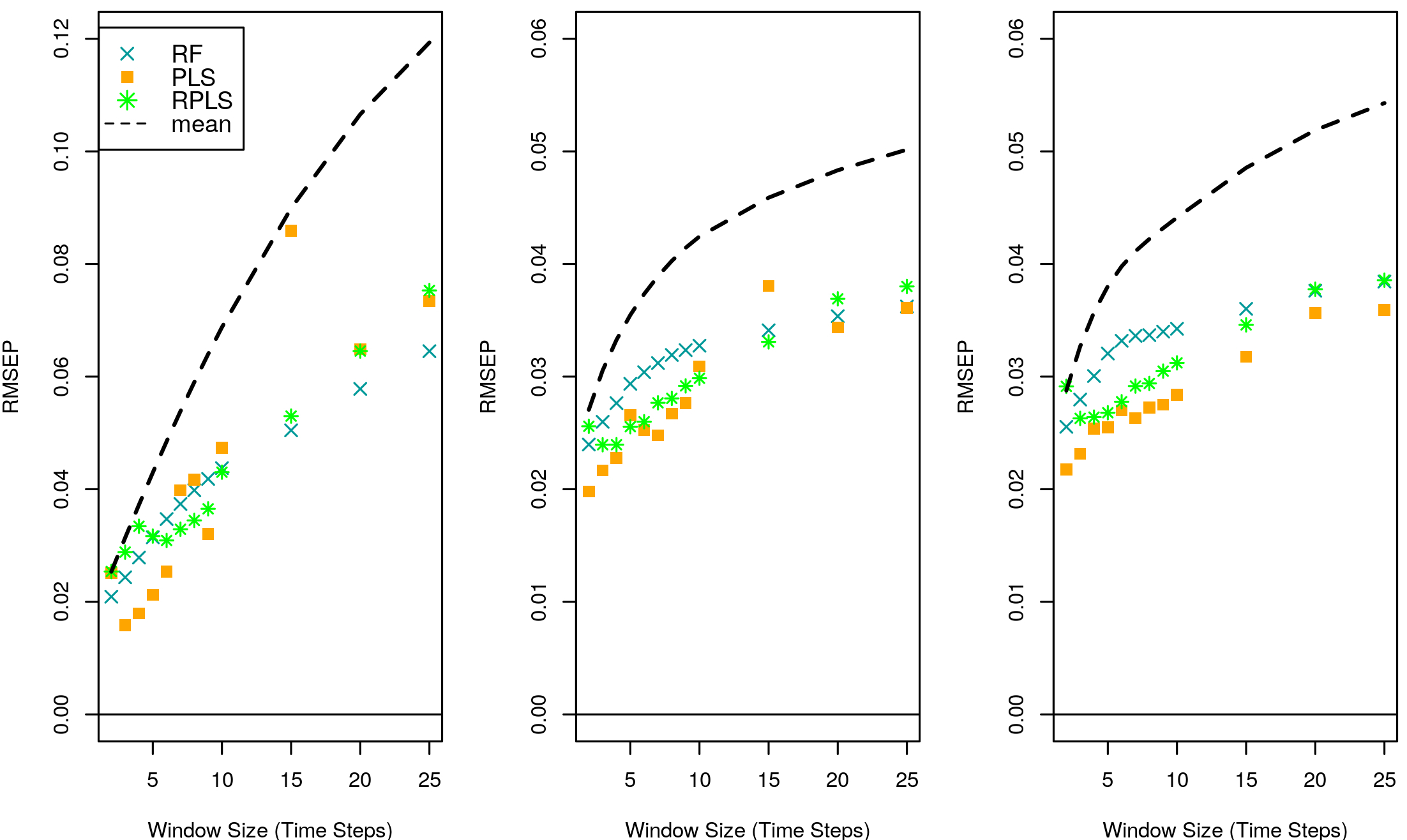}
	\caption{One step ahead RMSEP of the soft sensor models vs window size used in the calibration on the debutanizer (left), sulfur recovery unit H$_2$S (center), and the sulfur recovery unit SO$_2$ (right) datasets.}
\end{figure}

This relationship between window size and errors of prediction was unanticipated, and it stands at odds with that seen from typical multivariate calibrations done on time independent data. The effect of using small sample sizes to build traditional multivariate calibration models is well known. Typically, small training sets fail to span all sources of variation and therefore introduce biases; these biases lead to calibration models that are not robust to noise or background effects \cite{sundberg}. However, the prediction errors reported in this study are derived from many local moving window models, in that the predictions formed from this experiment were one sample ahead of their calibration windows. In three of the datasets, local changes in the time-dependent covariance appeared to be best modeled by the use of small window sizes rather than by relatively large ones. This can be explained by the idea that models with window sizes that span more time than a changes that occurs in the autocovariance can be expected to be less responsive than a model which spans smaller periods of time \cite{rollingwindow}.

The finding that models built with small calibration window sizes performed better than large ones on these datasets was the basis for the development of models used throughout the rest of the study. We selected a window size of 4 for the PLS, RF, and RPLS regression methods used in the remaining studies because models built on windows of that size performed well in train-test split validation. These windows did not tend to reduce the local PLS models to least squares regression as occurs when the optimal number of latent variables is restricted to the rank of the sensor data contained in the leave one out cross-validated sample window. This window size was also congruent with the primary goal of this study, in that we wanted to assess how well small process models performed as soft sensors.

Due to the large number of possible random forest sample windows and PLS inner model sizes, the model validation for the window sizes used was made via trial and error on the first 4.0\% of samples in the debutanizer data. The best hybrid RF-PLS regression model that was found on the debutanizer data was created from a window size of four samples with the inclusion of an unknown sample prediction made via the PLS inner model, to ensure fair comparison. The PLS model inside of the hybrid RF-PLS models utilized only the three samples prior to the unknown in a moving window fashion. The same RF-PLS model window sizes that were used on the debutanizer data were also applied to every other dataset after nonexhaustive train-test split validation. 

The root-mean-square errors of prediction for one time step ahead predictions are summarized in Table 1. For the debutanizer data, we found that an RF-PLS model had a one-step-ahead predictive error of 1.44\%. This result was competitive with that from a more complex local partial least-squares soft sensor with adaptive state partitioning (1.20\%) \cite{StateOfTheArt}, which was trained locally and updated dynamically. Similarly, we found that the hybrid RF-PLS modeling method had lower one-step-ahead errors of prediction than a least squares support vector machine model (H$_2$S = 3.97\%, SO$_2$ = 5.12\%) or a relevance vector machine (H$_2$S = 4.01\%, SO$_2$ = 5.11\%) model applied to the SRU process data \cite{RVM}. We should note that, the previous results from the literature process data and ours are not directly comparable. For example, in \cite{RVM} the authors made a local model from 1,680 samples before regressing on a selected set of 1,680 samples out of the total 10,080 samples available; the work reported here considered all 10,080 samples in a moving window fashion. Despite the fact that the study reported here and that in \cite{RVM} were not directly comparable, the errors of prediction obtained in this study are similar in magnitude to those from models trained under different circumstances. 

% & Debutanizer & SRU H$_2$S & SRU SO$_2$ & Penicillin \\
% latex table generated in R 3.4.1 by xtable 1.8-2 package
% Wed Aug 16 18:11:38 2017
\begin{table}[ht]
	\centering
	\begin{tabular}{ccccc}
		\hline
		& Debutanizer & SRU H$_2$S & SRU SO$_2$  \\
		& (RMSEP \%) & (RMSEP \%) & (RMSEP \%)  \\
		\hline
		RF & 2.43 & 2.60 & 2.79 \\ 
		PLS & \textit{1.58} & \textit{2.16} & \textit{2.30} \\ 
		RPLS & 2.90 & 2.39 & 2.62 \\ 
		RF-PLS & \textbf{1.44} & \textbf{1.91} & \textbf{2.06}  \\ %need to update the SRU preds to their latest values...
		MMW & 3.14 & 3.06 & 3.26  \\ 
		\hline
	\end{tabular}
	\caption{Root mean squared errors of prediction of all the methods studied on the debutanizer and sulfur recovery unit datasets as studied here for one-step-ahead predictions. Entries in bold indicate the lowest RMSEP for a given dataset, and those in italics were the second lowest. }
\end{table}

The hybrid RF-PLS method reported here had a lower one-step-ahead RMSEP than that of the moving window RF and that of conventional moving window PLS on all of the datasets studied, a result which indicated that the incorporation of PLS predictions into the RF model was beneficial. The RF-PLS method also featured the lowest prediction error for the property values of the debutanizer and SRU process data. 

Some insight related to the performance of the methods used in this study which incorporated $X$ information was obtained from comparison of the one-step-ahead predictions of these methods against those from the mean-moving-window method. No instances were observed where any of the 4 models in Table 1 had a greater one-step-ahead RMSEP than that obtained from the mean-moving-window. This finding suggested that all of the methods used information in the process sensor matrix $X$ to model the property value in $y$ better than the mean-moving-window model did using only $y$. This result was anticipated for the sulfur recovery unit process datasets, but not expected from the debutanizer data because the debutanizer data features large regions of monotonic trends in the property value relative to the small window sizes that were studied, as can be seen in the supplemental information. Random forest regression should predict poorly under these circumstances because random forest regression cannot be used to predict property values outside of its training set, as occurs during a monotonic increase. It appears that random forest regression can impart an upwards or downward bias in its predictions relative to the mean value of previous reference measurements. This bias may not model a process as accurately as partial least squares calibrations performed on small windows of the process data, but often the predicted one-step-ahead property values still closely resembled the actual value and produced relatively low predictive errors.

\subsection{Update Condition Studies}
While there is merit in being able to predict one time step ahead, and to do so with low error, we focused on predicting several time steps ahead. The laboratory analysis of the property value of a sample could be delayed by one or more sampling times \cite{SSBook}. Two cases for handling such events were investigated. The first case, a delayed update, is employed when the process model is not updated until a fixed number of new calibration samples becomes available. The second case, a continuous update, is employed when updates to the model are performed as soon as the next measured property value becomes available. The one-time-step ahead predictions discussed above can be seen as the degenerate solution to either updating a model at every single time step, or predicting ahead from a window by one time step. Both update conditions were investigated for the three process data sets, using different time delays before updating the calibration.

Figure 4 shows the errors of prediction that were obtained from the debutanizer and sulfur recovery unit data sets under the delayed update conditions. Within the range of one to nine time steps for model updates, the moving window PLS and RPLS modeling methods were observed to provide predictions worse than those obtained from the mean value modeling on more than one dataset. Both the RF and RF-PLS modeling methods were more robust to time delays before window updates than the other methods. Neither RF-PLS and RF models performed worse in prediction than the mean predictor. This finding suggests that the random forest predictions on these datasets was always better suited to prediction than a model which did not incorporate information from $X$. 

Relative to the random forest regression method, additional performance gains were achieved by including PLS predictions into the RF models by using RF-PLS. In all but one out of 27 update conditions on all datasets, predictions from RF-PLS hybrid models were better than those from RF models alone. RF-PLS models also produced lower delayed update RMSEP values than those from PLS on these datasets. This result further indicated that predictions from RF were improved by incorporating a moving window of samples from an inner PLS model. Overall, hybrid RF-PLS modeling tended to produce the lowest errors of prediction for these datasets under the delayed update condition.

\begin{figure}[H]
	\centering
	\includegraphics[width=0.99\linewidth]{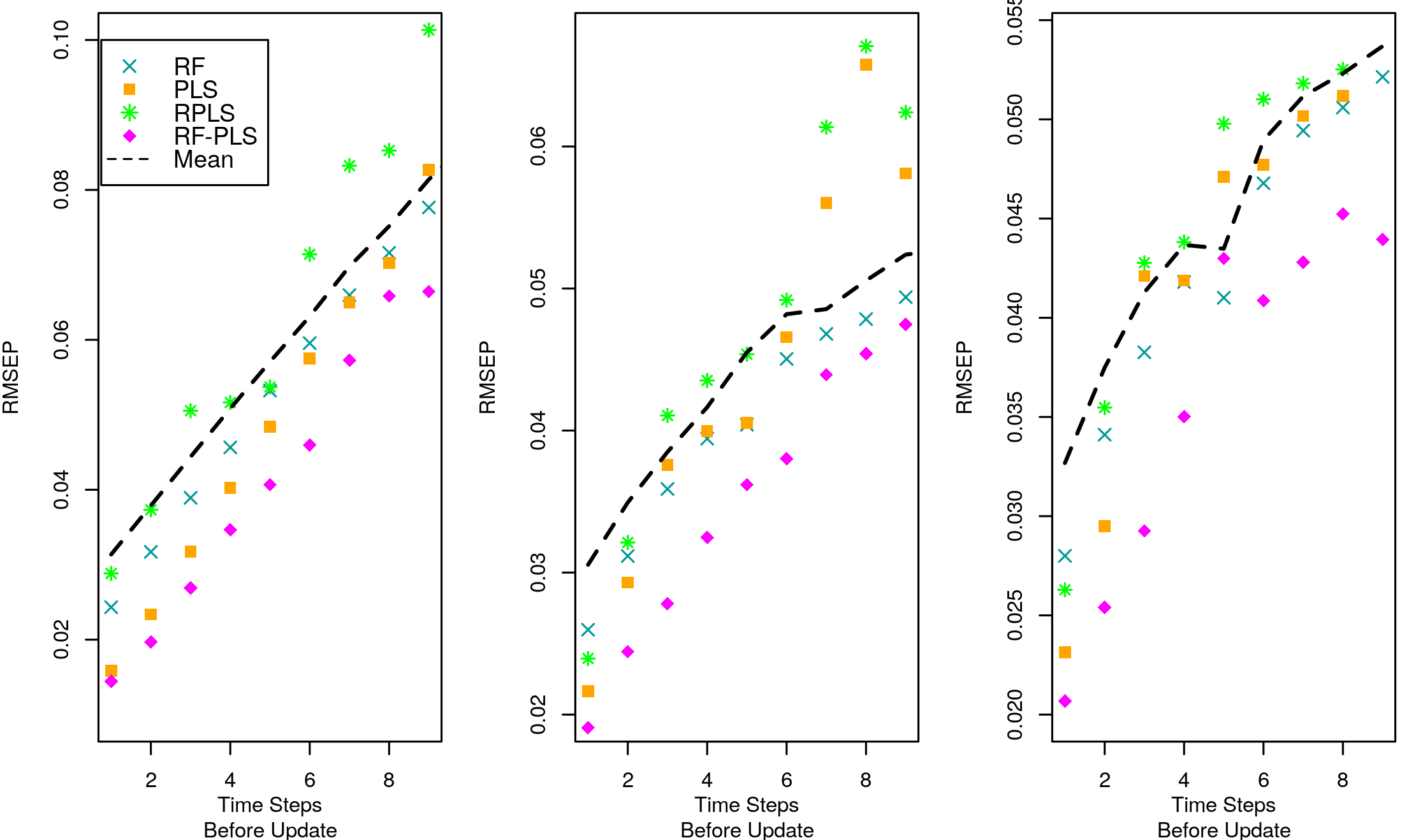}
	\caption{Model RMSEP as a function of delay for delayed model updates on the debutanizer (left), sulfur recovery unit H$_2$S (center), and on the sulfur recovery unit SO$_2$ (right) processes.}
\end{figure}

The continuously updated case assumes that a window of process measurement samples is always delayed from the sample requiring prediction by a fixed number of time steps. The continuously updated models tended to have greater predictive errors than those found using the same models when the delayed update case conditions were imposed, as can be seen in Figure 5. This result was expected, because the continuously updated models have a longer time delay from the window of currently available calibration samples to the samples requiring prediction than that of the delayed update. 

For example, if both update conditions are set to have an update of 4 sample steps, the model based on the continuous delay window is always predicting 4 samples ahead of its calibration window for each of the 4 samples, while the model based on the delayed update predicts 1, 2, 3 and 4 samples ahead of its calibration window for the same samples. On average, at time delays greater than one time step, the delayed update has more training samples closer to the samples it predicts upon, and can produce better predictions. 

\begin{figure}[H]
	\centering
	\includegraphics[width=0.99\linewidth]{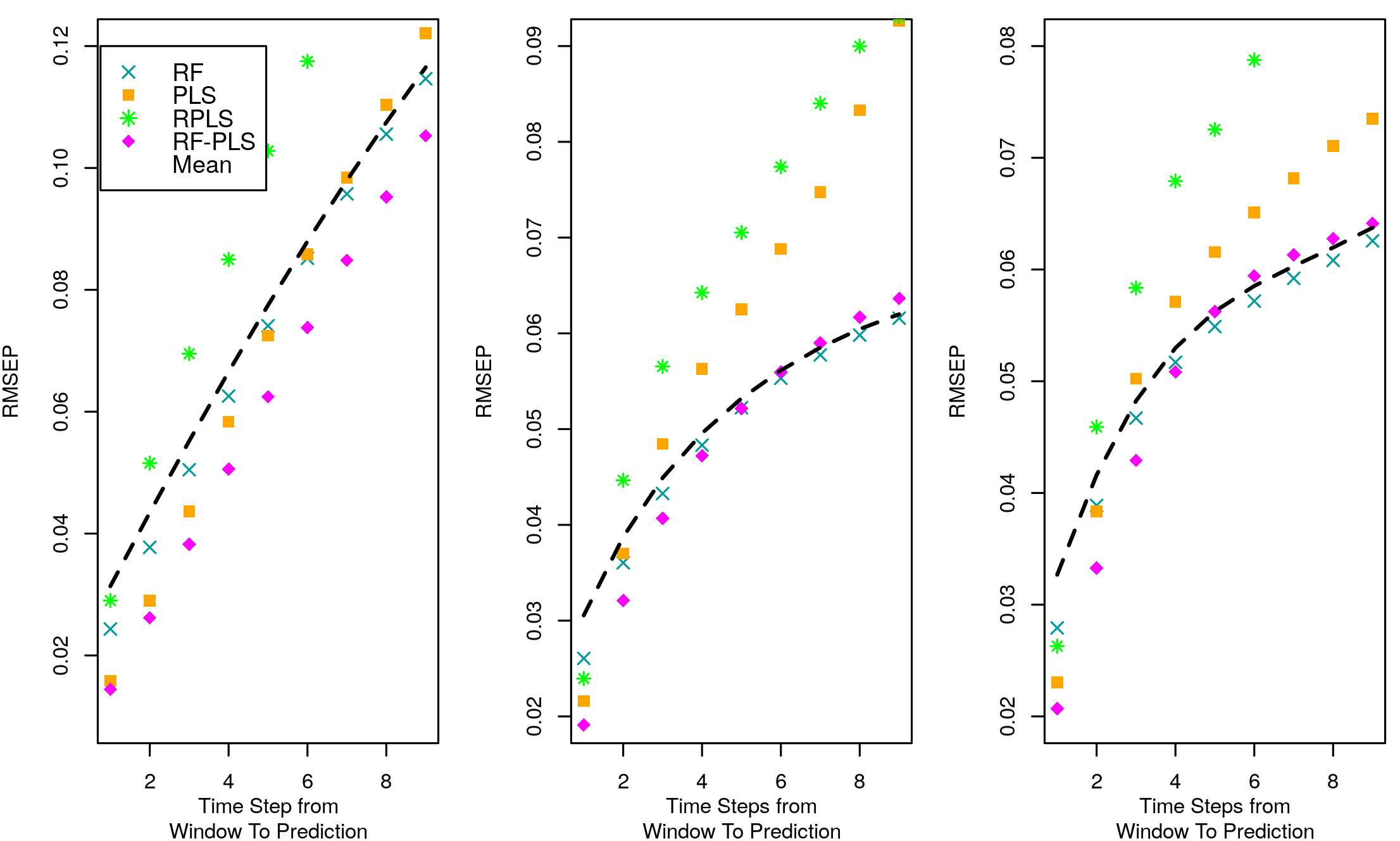}
	\caption{RMSEP as a function of time delay between the calibration window to a prediction for continuously updated models on the debutanizer (left), sulfur recovery unit H$_2$S (center), and on the sulfur recovery unit SO$_2$ (right) datasets.}
\end{figure}

For the continuously updated condition, only the random forest regression method produced better predictions than the MMW model in all datasets and over all time lags investigated. For the debutanizer data, the hybrid RF-PLS model afforded the lowest error of prediction for all nine time delays. However, when applied to the SRU process data, random forest partial least squares regression only generated the lowest error of prediction until the 4$^{th}$ time delay. When the model window was delayed 4 time steps from its prediction on the SRU process data, only RF regression modeling generated predictions better than those obtained from the mean-moving-window model. The most notable benefits from combining the PLS and RF predictions in the SRU dataset occurred at shorter time delays. However, the increase in root mean square error of prediction resulting from larger time delays (5-9) using RF-PLS hybrid modeling was much lower on the SRU data than that from either moving window PLS or RPLS modeling. 

The errors in prediction obtained from hybrid RF-PLS modeling most closely resembled those from RF modeling under the continuously updated condition. When the RF-PLS model failed to produce the best prediction results, these were never more than 3\% different than those from the mean-moving-window. The fact that errors from random forest partial least squares regression approached those obtained from the mean-moving-window predictor for time lags in situations where it failed to provide the lowest error of prediction is an interesting trait for a soft sensor. It is of potential interest because the failure trajectory of RF-PLS forecasting may be characterized by comparison with the mean-moving-window model. 

Overall, the random forest regression models performed better than the mean-moving-window model on these three datasets. In an industrial process where a moving window is applicable, and where the goal is to predict only a few samples ahead of the measurement of a property, conventional moving window partial least squares regression appears to be a good choice. However, combining PLS and RF modeling resulted in lower one-step-ahead predictions and in a majority of instances, lower predictions from the continuous update condition were observed than those from moving window partial least squares regression. Regardless of the updating conditions, for 97.77\% of the delay trials, the combination of PLS and RF in the hybrid RF-PLS model provided a lower predictive error than either moving window PLS or moving window RF alone. 

\subsection{Monotonic Property Value Data}
Although the random forest partial least squares ensemble modeling method showed promise in predicting all of the property values available in the debutanizer and the sulfur recovery unit process data, it performed less well on the penicillin data set. The simulated penicillin process data had one notable feature that distinguished it from the other datasets. Unlike the other property values studied here, the concentration of penicillin monotonically increased with time in this process, as can be seen in Figure 6 and in the supplementary information. The penicillin process data was used as a means to assess how well RF-PLS could model a monotonically increasing property value. It was expected that random forest regression models would perform poorly under monotonic changes in property values because random forest regression cannot extrapolate predictions outside of those which it was trained on. Therefore the penicillin data was a test case for the effect of introducing PLS predictions into an RF model, which otherwise had to predict outside of the calibration window.

\begin{figure}[H]
	\centering
	\includegraphics[width=0.75\linewidth]{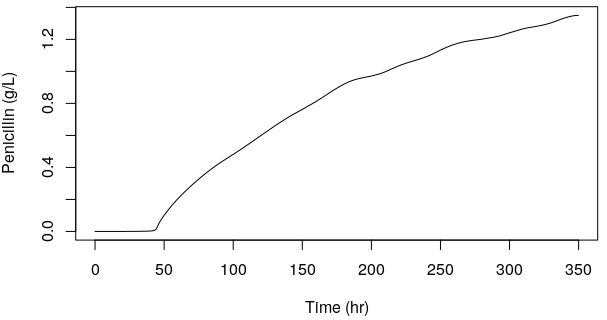}
	\caption{Simulated concentration of penicillin throughout the penicillin fermentation process.}
\end{figure}

The inversely proportional trend between window size and modeling accuracy that was observed on the debutanizer and SRU datasets was not seen with partial least squares regression and recursive PLS when these were applied to the simulated penicillin dataset. As seen in Figure 7, with the penicillin data, use of larger window sizes tended to produce PLS-based models with low errors of prediction. Similarly, at small window lengths, RPLS exhibited the lowest error in one-step-ahead predictions when compared to all of the other methods likely because of the recursive memory. The success of PLS methods for the simulated penicillin fermentation data is most likely attributable to small or infrequent changes in the $X$ covariance structure. The nearly uniform error produced by moving window RF modeling on this process data is explained by the fact that random forest regression cannot be used to predict values beyond the training range. The predictions made from the random forest regression moving window models were found to always be lower than the true property value most likely  because the property values were monotonically increasing. 

\begin{figure}[H]
	\centering
	\includegraphics[width=0.95\linewidth]{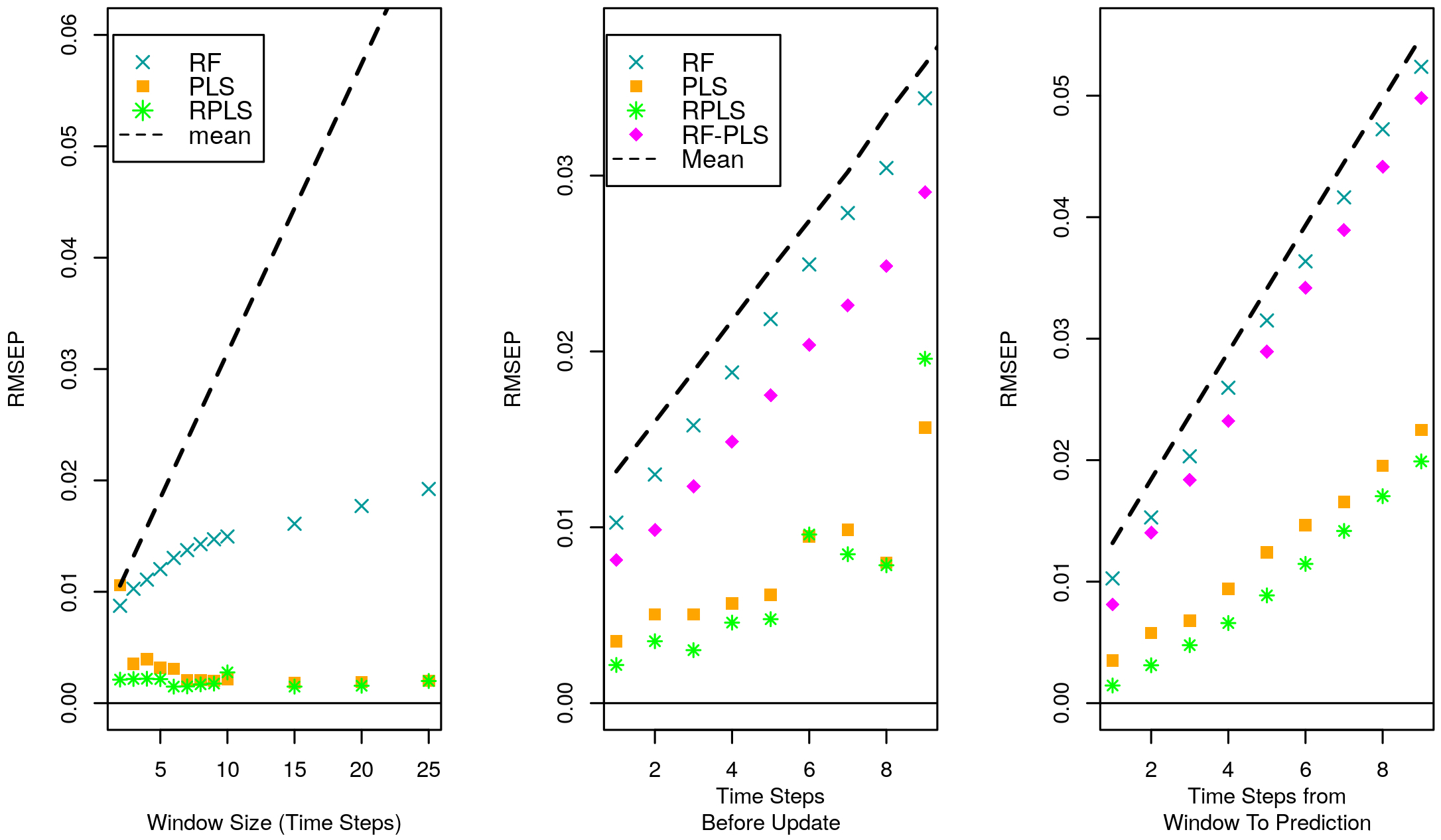}
	\caption{Experimental trials for the penicillin fermentation process data: error of prediction vs window size (left), RMSEP vs model update delay for the delayed update (center), RMSEP vs time step from calibration window to prediction for continuously updated models (right). }
\end{figure}

Both updating conditions, the continuous and delayed updates, were examined on the penicillin fermentation dataset. Both the RPLS and PLS modeling methods provided the lowest error for both model update cases at all delays studied. Again, the random forest modeling method could not be used to predict values outside of those from the process data contained in its training window, and had systematic, negatively-biased predictions. Although the coupling of a PLS prediction with the calibration window for the RF-PLS method produced uniformly lower errors of prediction than those obtained through the use of RF regression alone, RF-PLS modeling also did not perform as well as either of the other PLS methods. For datasets similar to the penicillin fermentation data, one could expect that moving window PLS based methods would be better suited. 

Different trends were observed with relation to the modeling error and time delays before model updates/prediction were observed for the other data sets, none of which had monotonic $y$ structures. While the RF-PLS modeling technique could be expected to perform poorly on the penicillin process data the predictions from RF-PLS modeling still outperformed those from random forest regression in both window update conditions, and with respect to one-step-ahead errors of prediction, as shown in Table 2.

\begin{table}[ht]
	\centering
	\begin{tabular}{ccccccc}
		& RF & PLS & RPLS & RF-PLS & Mean \\
		\hline
		RMSEP (g/L) & 0.0102 & \textit{0.0035} & \textbf{0.0014} & 0.0081 & 0.0131 \\
		\hline
	\end{tabular}
	\caption{Root mean squared errors of prediction on the pencillin dataset for all the methods studied using one-step-ahead predictions. The entry in bold indicates the lowest RMSEP for a given technique, and the one in italics indicates the second lowest. }
\end{table}

\section{Conclusion}
Although the nature of the data was an important factor in the success of the models studied, so too were the model update conditions and the calibration window sizes used. We found that by using small calibration window sizes, PLS, RPLS, and RF regression models offered low one-step-ahead predictive errors relative to those obtained by using larger window sizes. The one-step-ahead errors obtained with these simple methods that used very few samples were lower or similar in magnitude to the prediction errors reported on the sulfur recovery unit data set, and on the debutanizer process data which used different calibration techniques. The memory mechanisms that were a part of recursively updated, small calibration window models, such as RPLS, tended to produce poorer predictions on data than those produced by rebuilding the model on a small window of samples for two of the datasets under continuously updated conditions. 

The new soft-sensing method, RF-PLS, was the most robust to time delay between the calibration window and its unknown sample while still generating the lowest error for one-step-ahead predictions on three of the datasets reported here. Its application to modeling processes which have monotonic property values appears limited, much like moving window random forest regression. On process property values that were not entirely monotonic, such as the debutanizer or sulfur recovery unit data, the RF-PLS models offered the lowest predictive errors for the delayed update case, and in most instances, for the continuously updated case. On datasets that did not have monotonic trends in $y$, the combination of random forest regression and partial least squares was advantageous in the majority of update conditions and led to lower predictive errors than either method separately.

\section{Acknowledgments}
This work was supported by the United States National Science Foundation grant 1506853.

\section{Conflict of Interest}
The authors declare no conflict of interest.

\section{References}


%% References


\bibliographystyle{model1-num-names}
\bibliography{sample.bib}



\newpage
\section{Supplemental Information}

\begin{figure}[H]
	\centering
	\includegraphics[width=0.8\linewidth]{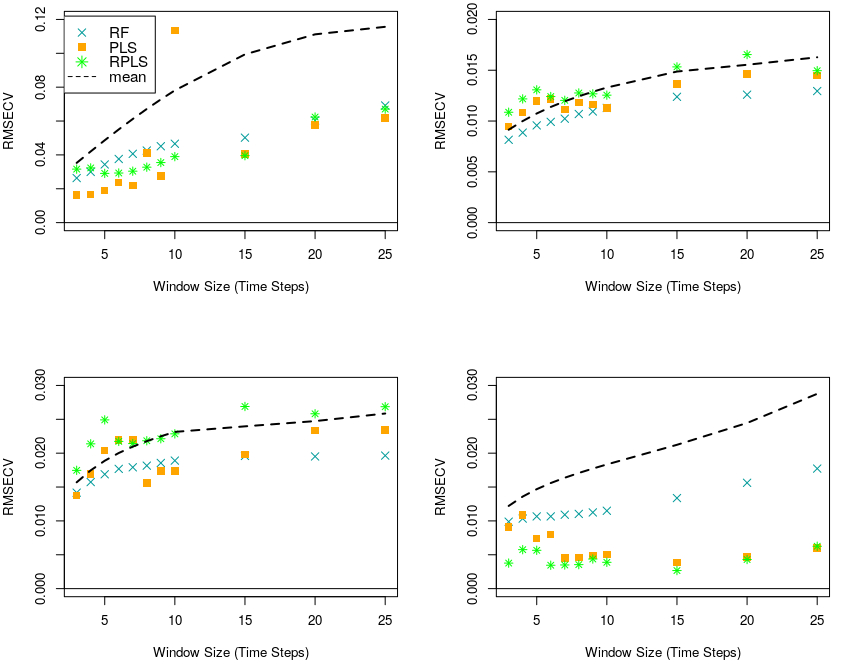}
	\caption{Train-test split validation of window size for the PLS, RF, and RPLS moving window methods on the debutanizer (top-left), sulfur recovery unit H$_2$S (top-right), sulfur recovery unit SO$_2$ (bottom-left), and penicillin data sets (bottom-right).}
	\includegraphics[width=0.8\linewidth]{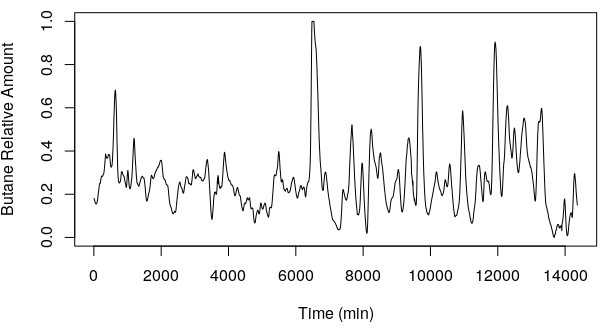}
	\caption{Relative amount of butane throughout the debutanizer process.}
\end{figure}

\begin{figure}[!h]
	\centering
		\includegraphics[width=0.8\linewidth]{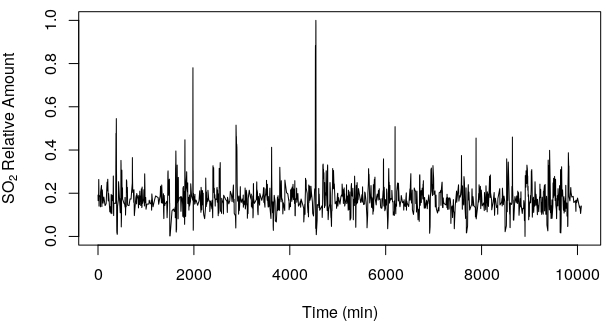}
	\caption{Relative amount of SO$_2$ throughout the sulfur recovery unit process.}
	\includegraphics[width=0.8\linewidth]{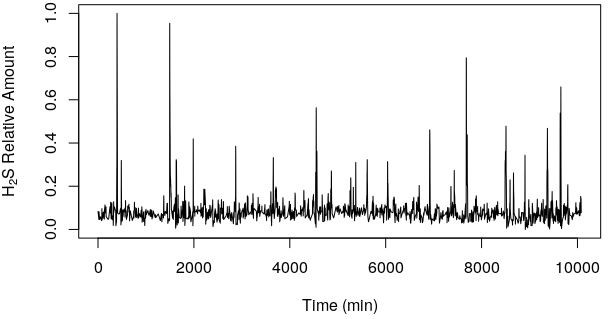}
	\caption{Relative amount of H$_2$S throughout the sulfur recovery unit process.}
\end{figure}

\end{document}